\documentclass[10pt,twocolumn,letterpaper]{article}

\usepackage{cvpr}
\usepackage{times}
\usepackage{epsfig}
\usepackage{graphicx}
\usepackage{amsmath}
\usepackage{amssymb}
\usepackage{authblk}
\usepackage{svg}
\usepackage{multirow}
\usepackage{caption}
\usepackage[pagebackref=true,breaklinks=true,letterpaper=true,colorlinks,bookmarks=false]{hyperref}

\cvprfinalcopy 

\def\cvprPaperID{6954} 

\usepackage{nopageno}

\ifcvprfinal\fi
\begin{document}

\title{Multi-Agent Tensor Fusion for Contextual Trajectory Prediction}

\makeatletter
\renewcommand\AB@affilsepx{, \protect\Affilfont}
\makeatother

\renewcommand*{\Authsep}{, }
\renewcommand*{\Authand}{, }
\renewcommand*{\Authands}{, }

\author[2,3]{Tianyang Zhao}
\author[1,3]{Yifei Xu}
\author[1,4]{Mathew Monfort}
\author[1]{Wongun Choi}
\author[1]{Chris Baker}
\author[1]{Yibiao Zhao }
\author[2]{Yizhou Wang}
\author[1,3]{Ying Nian Wu}

\renewcommand\Affilfont{\fontsize{9}{10.8}\itshape}
\affil[1]{\href{http://isee.ai}{ISEE.AI}}
\affil[2]{Peking University}
\affil[3]{UCLA}
\affil[4]{MIT CSAIL
\authorcr \{zhaotianyang, yizhou.wang\}@pku.edu.cn, \{mmonfort, wchoi, chrisbaker, yz\}@isee.ai, fei960922@ucla.edu, ywu@stat.ucla.edu}

\maketitle

\begin{abstract}

Accurate prediction of others' trajectories is essential for autonomous driving. Trajectory prediction is challenging because it requires reasoning about agents' past movements, social interactions among varying numbers and kinds of agents, constraints from the scene context, and the stochasticity of human behavior. Our approach models these interactions and constraints jointly within a novel Multi-Agent Tensor Fusion (MATF) network. Specifically, the model encodes multiple agents' past trajectories and the scene context into a Multi-Agent Tensor, then applies convolutional fusion to capture multiagent interactions while retaining the spatial structure of agents and the scene context. The model decodes recurrently to multiple agents' future trajectories, using adversarial loss to learn stochastic predictions. Experiments on both highway driving and pedestrian crowd datasets show that the model achieves state-of-the-art prediction accuracy.

\end{abstract}

\section{Introduction}

Human drivers continually anticipate the behavior of nearby vehicles and pedestrians in order to plan safe and comfortable interactive motions that avoid conflict with others.
Autonomous vehicles (AVs) must likewise predict the trajectories of others in order to proactively plan for future interactions \emph{before} they occur, rather than reactively respond to unanticipated outcomes \emph{after} they occur, which can lead to unsafe behaviors such as sudden hard braking, or failure to execute maneuvers in dense traffic. Fundamentally, trajectory prediction allows autonomous vehicles to reason about the possible future situations they will encounter, to evaluate the risk of a given plan relative to these predicted situations, and to select a plan which minimizes that risk.
This adds a layer of interpretability to the system that is critical for debugging and verification.

\begin{figure}[t!]
\begin{center}
    \includegraphics[width=0.85\linewidth]{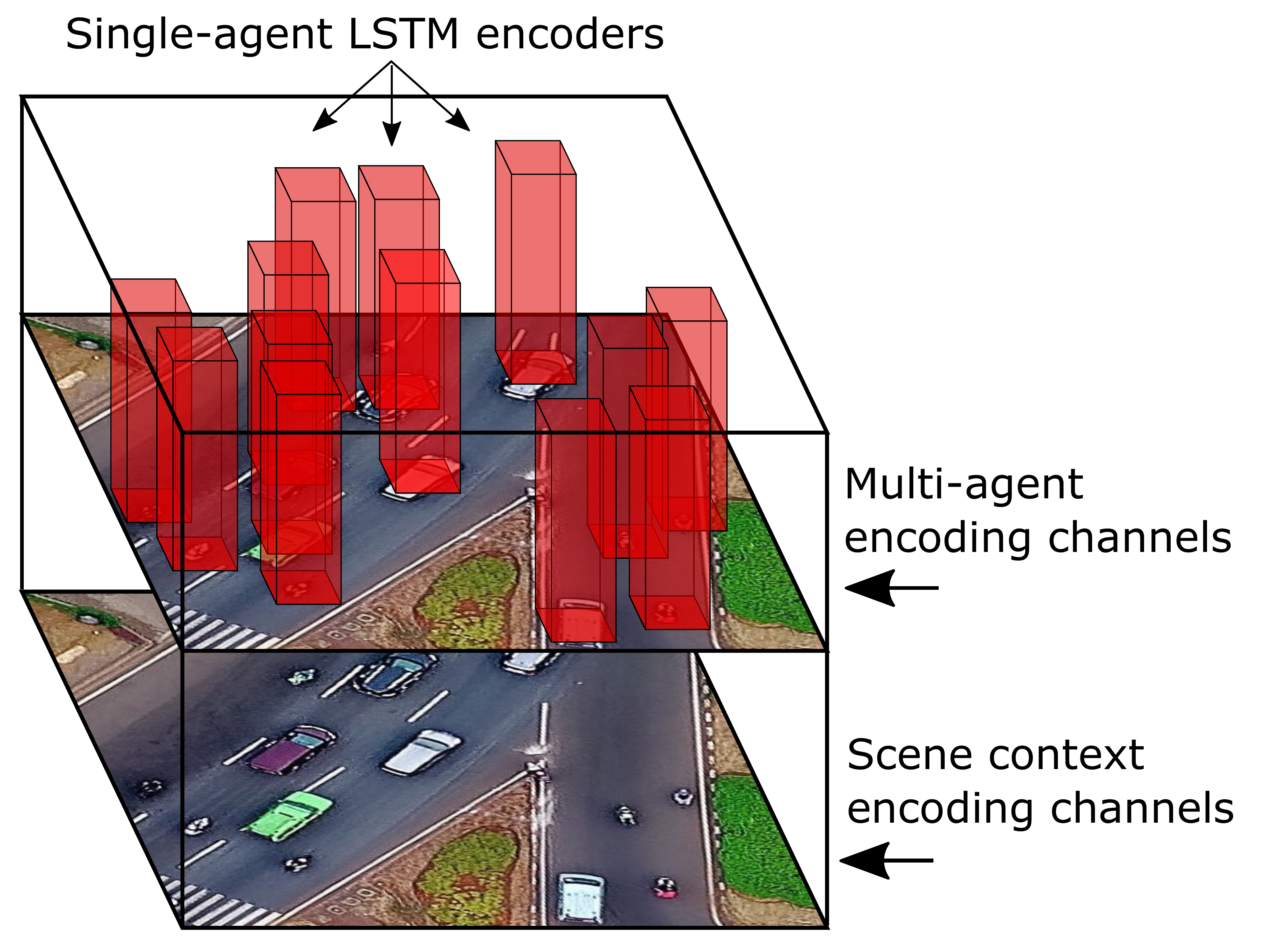}
\end{center}
   \caption{The Multi-Agent Tensor encoding is a spatial feature map of the scene context and multiple agents from an overhead perspective, including agent channels (above) and context channels (below). Agents' feature vectors (red) output from single-Agent LSTM encoders are placed spatially w.r.t. agents' coordinates to form the agent channels. The agent channels are aligned spatially with the context channels (a context feature map) output from scene context encoding layers to retain the spatial structure.}
\label{fig:short}
\end{figure}

Trajectory prediction is challenging because agents' motions are stochastic, and dependent on their goals, social interactions with other agents, and the scene context. Predictions must generalize to new situations, where the number and configuration of other agents are not fixed in advance. Encoding this information is difficult for neural-network-based approaches, because standard NN architectures prefer fixed input, output, and parameter dimensions, while for the prediction task these dimensions vary. Previous work has addressed this issue using either agent-centric or spatial-centric encodings. Agent-centric encodings apply aggregation functions on multiple agents' feature vectors, while spatial-centric approaches operate directly on top-down representations of the scene.

We propose a novel Multi-Agent Tensor Fusion (MATF) encoder-decoder architecture, which combines the strengths of agent- and spatial-centric approaches within a flexible network that can be trained in an end-to-end fashion to represent all the relevant information about the social and scene context. The Multi-Agent Tensor representation, illustrated in Fig. 1, spatially aligns an encoding of the scene with encodings of the past trajectory of every agent in the scene, which maintains the spatial relationships between agents and scene features. Next, a fused Multi-Agent Tensor encoding is formed via a fully convolutional mapping (see Fig. 2), which naturally learns to capture the spatial locality of interactions between multiple agents and the environment, as in agent-centric approaches, and preserves the spatial layout of all agents within the fused Multi-Agent Tensor in a spatial-centric manner.

Our model decodes the comprehensive social and contextual information encoded by the fused Multi-Agent Tensor into predictions of the trajectories of all agents in the scene simultaneously. Real-world behavior is not deterministic -- agents can perform multiple maneuvers from the same context (e.g. follow lane or change lane), and the same maneuver can vary in execution in terms of velocity and orientation profile. We use conditional generative adversarial training~\cite{gan, cgan} to capture this uncertainty over predicted trajectories, representing the distribution over trajectories with a finite set of samples.

We conduct experiments on both driving datasets and pedestrian crowd datasets. Experimental results are reported on the publicly available NGSIM driving dataset~\cite{ngsim}, Stanford Drone pedestrian crowd dataset~\cite{stanford_drone}, ETH-UCY crowd datasets~\cite{ucy, eth}, and a private recently-collected Massachusetts driving dataset. Quantitative and qualitative ablative experiments are conducted to show the contribution of each part of the model, and quantitative comparisons with recent approaches show that the proposed approach achieves state-of-the-art accuracy in both highway driving and pedestrian trajectory prediction.

\section{Related Work}

Traditional methods for predicting or classifying trajectories model various kinds of interactions and constraints by hand-crafted features or cost functions~\cite{interaction_maneuver, track, collective, vehicle, social_force, detection_social_force, who_are_you}.
Early methods based on inverse optimal control also use hand-crafted cost features, and learn linear weighting functions to rationalize trajectories which are assumed to be generated by optimal control~\cite{oc}.
Recent data-driven approaches based on deep networks~\cite{social_lstm, chauffeur, conv_social_pooling, maneuver, social_gan, gail_gru, desire, r2p2, sophie, carnet, social_attention} outperform traditional approaches. Most of this work focuses either on modeling constraints from the scene context~\cite{carnet} or on modeling social interactions among multiple agents~\cite{social_lstm, conv_social_pooling, maneuver, social_gan, social_attention}; a smaller fraction of work considers both aspects~\cite{chauffeur, desire, sophie}.

Agent-centric NN-based approaches integrate information from multiple agents by applying aggregation functions on multiple agents' feature vectors output from recurrent units. Social LSTM~\cite{social_lstm} runs max pooling over state vectors of nearby agents within a predefined distance range, but does not model social interaction with far-away agents. Social GAN~\cite{social_gan} contributes a new pooling mechanism over all the agents involved in a scene globally, and by using adversarial training to learn a stochastic, generative model of human behavior~\cite{gan}. Although these kinds of max pooling aggregation functions handle varying numbers of agents well, permutation invariant functions may discard information when input agents lose their uniqueness~\cite{sophie}. In contrast, Social Attention~\cite{social_attention} and Sophie~\cite{sophie} address the heterogeneity of social interaction among different agents by attention mechanisms~\cite{attention, only_attention}, and spatial-temporal graphs~\cite{srnn}. Attention mechanisms encode which other agents are most important to focus on when predicting the trajectory of a given agent. However, attention-based approaches are very sensitive to the number of agents included --- predicting $n$ agents has $O(n^2)$ computational complexity. In contrast, our approach captures multiagent interactions while maintaining $O(n)$ computational complexity.

The agent-centric approaches discussed above do not make use of spatial relationships among agents directly. As an alternative, spatial-centric approaches retain the spatial structure of agents and the scene context throughout their representations. Convolutional Social Pooling~\cite{conv_social_pooling} partially retains the spatial structure of agents' locations by forming a social tensor which is similar to our Multi-Agent Tensor representation, but much of this spatial information is later aggregated by several bottleneck layers. This approach does not encode the scene context, and only a single agent's trajectory can be predicted with each forward pass --- potentially too slow for real-time trajectory prediction of multiple agents. Chauffeur Net~\cite{chauffeur} proposes a novel method to retain the spatial structure of agents and the scene by directly operating on the spatial feature map of agents and the scene context. In this approach, agents are represented as bounding boxes and do not have independent recurrent encoding units. In contrast, our model encodes multiple agents' feature vectors via recurrent units while simultaneously retaining the spatial structure of agents and the scene throughout the reasoning process.

Many data-driven approaches learn to predict deterministic future trajectories of agents by minimizing reconstruction loss~\cite{social_lstm, carnet}. However, human behavior is inherently stochastic. Recent approaches address this by predicting a distribution over future trajectories by combining Variational Auto-Encoders~\cite{vae} and Inverse Optimal Control~\cite{desire}, or with conditional Generative Adversarial Nets~\cite{social_gan, sophie}. GAIL-GRU~\cite{gail_gru} uses generative adversarial imitation learning~\cite{gail} to learn a stochastic policy that reproduces human expert driving behavior. R2P2~\cite{r2p2} proposes a novel cost function to encourage enhancement in both precision and diversity of the learned predictive distribution. Other approaches predict a set of possible trajectories, instead of a single deterministic trajectory, by conditioning on possible maneuver classes~\cite{conv_social_pooling, maneuver}.

\section{Method}

In this section, we describe the Multi-Agent Tensor Fusion (MATF) encoder, and the decoder architecture for trajectory prediction. The network is shown in Fig.~2. The network takes as input 1) the past trajectories of multiple dynamic interacting agents, and 2) a scene containing a static context, which is represented from an overhead perspective and can either be a segmented image containing all static objects, or a bird's-eye view raw image. The network outputs the predicted future trajectories of all agents in the scene.

\begin{figure}[h!]
\begin{center}
    \includegraphics[width=1.0\linewidth]{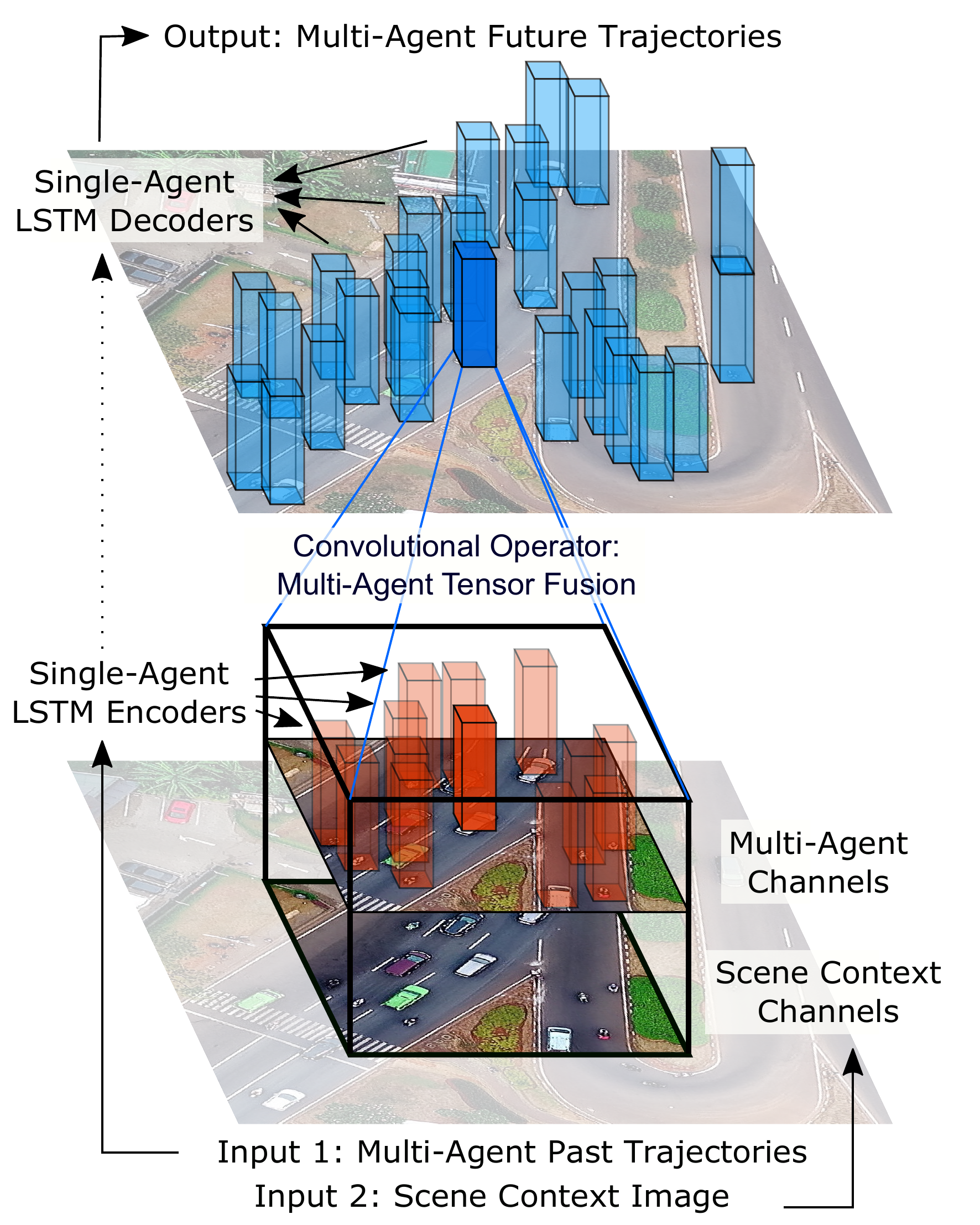}
\end{center}
   \caption{Illustration of the Multi-Agent Tensor Fusion (MATF) architecture. The inputs are $n$ agent past trajectories $\{ x_1, x_2, .. , x_n \}$ over a past time period of length $T$, and a bird's-eye view scene context image $c$. Each $x_i$ and $c$ are encoded independently through recurrent and convolutional encoding streams, respectively. The encoded agent vectors $\{ x_1', x_2', .. , x_n' \}$ and encoded scene context $c'$ are aligned spatially together to form a Multi-Agent Tensor. The 3-D box shows a slice of the tensor surrounding the orange agent. Next, U-Net like fully convolutional spatial fusion layers are applied on top of the Multi-Agent Tensor to reason about the interactions while retaining spatial locality to output a fused Multi-Agent Tensor $c''$. 
   Fused vectors for each agent $\{ x_1'', x_2'', .. , x_n'' \}$ (blue boxes), are then sliced out from $c''$, and contain the interaction, history, and constraint features for the corresponding agent. Note that because we run shared convolution, the corresponding fused vectors of all the agents are generated in one forward pass. For instance, $x_i''$ (see solid blue box), contains fused information from all the agents and scene features near agent $i$ within the receptive field of the convolutional layers. These fused vectors are then added to the original encoded vectors of the corresponding agents as residuals to obtain final agent encoding vectors $x_i' + x_i''$, which are decoded independently to future trajectory predictions $\hat{y}_i$ over a future period of length $T'$. The whole architecture is fully differentiable and trained end-to-end.}
\label{fig:short}
\end{figure}

\subsection{MATF Encoding}

There are two parallel encoding streams in the MATF architecture. One encodes the past trajectories of each individual agent $x_i$ independently using single agent LSTM encoders, and another encodes the static scene context image $c$ with a CNN. Each LSTM encoder shares the same set of parameters, so the architecture is invariant to the number of agents in the scene. The outputs of the LSTM encoders are 1-D agent state vectors $\{ x_1', x_2' , .. , x_n' \}$ without temporal structure. The output of the scene context encoder CNN is a scaled feature map $c'$ retaining the spatial structure of the bird's-eye view static scene context image.

Next, the two encoding streams are concatenated spatially into a Multi-Agent Tensor. Agent encodings $\{ x_1', x_2' , .. , x_n' \}$ are placed into one bird's-eye view spatial tensor, which is initialized to 0 and is of the same shape (width and height) as the encoded scene image $c'$. The dimension axis of the encodings fits into the channel axis of the tensor as shown in Fig.~1. The agent encodings are placed into the spatial tensor with respect to their positions at the last time step of their past trajectories. This tensor is then concatenated with the encoded scene image in the channel dimension to get a combined tensor. If multiple agents are placed into the same cell in the tensor due to discretization, element-wise max pooling is performed.

The Multi-Agent Tensor is fed into fully convolutional layers, which learn to represent interactions among multiple agents and between agents and the scene context, while retaining spatial locality, to produce a fused Multi-Agent Tensor. Specifically, these layers operate at multiple spatial resolution scale levels by adopting U-Net-like architectures~\cite{unet} to model interaction at different spatial scales. The output feature map of this fused model $c''$ has exactly the same shape as $c'$ in width and height to retain the spatial structure of the encoding.

\subsection{MATF Decoding}

To decode each agent's predicted trajectory, agent-specific representations with fused interaction features for each agent $\{ x_1'', x_2'', .. , x_n'' \}$ are sliced out according to their coordinates from the fused Multi-Agent Tensor output $c''$ (Fig.~2). These agent-specific representations are then added as a residual~\cite{resnet} to the original encoded agent vectors to form final agent encoding vectors $\{ x_1'+x_1'',x_2'+ x_2'', ... , x_n'+x_n'' \}$, which encode all the information from the past trajectories of the agents themselves, the static scene context, and the interaction features among multiple agents. In this way, our approach allows each agent to get a different social and contextual embedding focused on itself. Importantly, the model gets these embeddings for multiple agents using shared feature extractors instead of operating $n$ times for $n$ agents.

Finally, for each agent in the scene, its final vector $x_i'+x_i''$ is decoded to future trajectory prediction $\hat y_i$ by LSTM decoders. Similar to the encoders for each agent, parameters are shared to guarantee that the network can generalize well when the number of agents in the scene varies.

The whole architecture is fully differentiable and can be trained end-to-end to minimize reconstruction loss between predicted future trajectories $\{ \hat y_1, \hat y_2, .. , \hat y_n \}$ and observed ground-truth future trajectories $\{y_1, y_2, .. , y_n \}$: $L_{L2/L1}(\hat{y}_i,y_i) = \sum_{t = 1} ^ {T'} L2/L1(\hat{y}_{it},y_{it})$, where $L2/L1$ indicates that we can use either the $L2$ or $L1$ distance between two positions for  reconstruction error.

\subsection{Adversarial Loss}

We use conditional generative adversarial training \cite{gan, cgan} to learn a stochastic generative model that captures the multimodal uncertainty of our predictions. GANs consist of two networks, a generator $G$ and a discriminator $D$ competing against each other. $G$ learns the distribution of the data and generates samples, while $D$ learns to distinguish the feasibility or infeasibility of the generated samples. These networks are simultaneously trained in a two player min-max game framework.

In our setting, we use a conditional $G$ to generate future trajectories of multiple agents, conditioning on all the agents' past trajectories, the static scene context, and random noise input to create stochastic outputs. Simultaneously, we use $D$ to distinguish whether the generated trajectories are real (ground truth) or fake (generated). Both $G$ and $D$ share exactly the same architecture in their encoding parts with the deterministic model presented in Section 3.1, to reason about static scene context and interaction among multiple agents spatially. Both $G$ and $D$ are initialized with parameters from the trained deterministic model introduced in previous subsections. Detailed architectures and losses are described below.

\textbf{Generator (G)} $G$ observes past trajectories of all the agents in a given scene $\{ x_1, x_2, .. , x_n \}$, and the static scene context $c$. It jointly outputs the predicted future trajectories $\{ \hat y_1, \hat y_2, .. , \hat y_n \}$ by decoding the final agent vectors $\{ x_1'+x_1'', x_2'+x_2'', ... , x_n'+x_n'' \}$ described in Section 3.2, concatenated with Gaussian white noise vector $z$. The architecture is exactly the same as presented in previous subsections, except that in the deterministic model, the final encoding for a given agent $x_i'+x_i''$ is concatenated with $z = 0$ vector to decode into its future trajectory; while in $G$, $z$ is sampled from a Gaussian distribution.

\textbf{Discriminator (D)} $D$ observes the ground truth past trajectories of all the agents in a given static scene context, combined either with all generated future trajectories $\{ x_1, x_2, .. , x_n,  \hat y_1, \hat y_2, ... , \hat y_n  \}$ or all ground truth future trajectories $\{ x_1, x_2, .. , x_n, y_1, y_2, .. , y_n \}$.
It outputs real or fake labels for the future trajectory of each agent in the scene, such that $D(y) = 0$ if trajectory $y$ is fake, and $D(y) = 1$ if trajectory $y$ is real.
$D$ shares nearly the same architecture as presented in previous subsections, except for the following differences: (1) Its single agent LSTM encoders take in past and future trajectories as input instead of just past trajectories; (2) As a classifier, it does not use an LSTM to decode the final agent vector $x_i'+x_i''$ to a future trajectory. Instead, final agent encodings are fed into fully connected layers to be classified as real or a fake.

\textbf{Losses} The adversarial loss $L _ {GAN}$ for a given scene is:
\begin{multline}
\mathcal L_{GAN}(scene) =  \\
\min_G \max_D \sum _ {i \in scene} \log D(y_i) + \log (1 - D(\hat y_i)),
\end{multline}
where $\{i|i\in scene\}$ is the set of agents in a given scene, $y_i$ and $\hat y_i$ denote ground truth (real) and generated (fake) trajectories, respectively, and $G$ denotes the generative MATF network which we are optimizing.

To train the MATF GAN, we use the following losses:
\begin{multline}
\Theta^* = \operatorname*{arg\,max}_{\Theta} \mathbb E_{scene} [\mathcal L _{GAN}(scene) \\
+ \lambda \mathcal \sum _ {i \in scene} L _ {L2 / L1} (\hat y_i, y_i)],
\end{multline}
where $\Theta$ is the set of parameters of the model and $\lambda$ weights the contribution of reconstruction loss versus adversarial loss.

\section{Experiments}

In the Experiments and Results sections, we evaluate our model on both driving datasets~\cite{ngsim} and pedestrian crowd datasets~\cite{ucy, eth, stanford_drone}. We construct different baseline variants of our models for ablative studies, and compare with state-of-the-art alternative methods quantitatively~\cite{social_lstm, vehicle, conv_social_pooling, social_gan, social_force, gail_gru, desire, sophie}. Qualitative results are also presented for further analysis.

\subsection{Datasets}

We use the publicly available NGSIM dataset \cite{ngsim}, a recently collected Massachusetts driving dataset, the publicly available ETH-UCY datasets~\cite{ucy, eth}, and the publicly available Stanford Drone dataset \cite{stanford_drone} for training and evaluation.

\textbf{NGSIM}. A driving dataset consisting of trajectories of real freeway traffic over a time span of 45 minutes. Data were recorded by fixed bird's-eye view cameras placed over a 640-meter span of US101. Trajectories of all the vehicles traveling through the area within this 45 minutes are annotated. The dataset consists of various traffic conditions (mild, moderate and congested), and contains around 6k vehicles in total.

\textbf{ETH-UCY}. A collection of relatively small benchmark pedestrian crowd datasets. There are 5 datasets with 4 different scenes, including 1.5k pedestrian trajectories in total. We use the same cross-validation training-test split metrics as reported in previous work~\cite{social_gan, sophie}.

\textbf{Stanford Drone}. A large-scale pedestrian crowd dataset consisting of 20 unique scenes in which pedestrians, bicyclists, skateboarders, carts, cars, and buses navigate on a university campus. Raw, static scene context images are provided from bird's-eye view, and coordinates of multiple agents' trajectories are provided in pixels. These scenes contain rich human-human interactions, often taking place within high density crowds, and diverse physical landmarks such as buildings and roundabouts that must be avoided. We use the standard test set for quantitative evaluation. Some scenes from the standard training set are not used for our training process, but left out for qualitative evaluation instead.

\subsection{Baseline Models}

We construct a set of baseline variants of our model for ablative studies.

\textbf{LSTM}: A simple deterministic LSTM encoder-decoder. It shares exactly the same architecture as the single-agent LSTM encoders and decoders introduced in Section 3 for fair comparison.

\textbf{Single Agent Scene}: This deterministic model shares exactly the same architecture as introduced in Section 3, except that it only takes in one agent history $x_i$ with scene representation $c$ and outputs only $\hat y_i$ each time, so the model reasons about scene-agent interaction, but is completely unaware of multi-agent interaction.

\textbf{Multi Agent}: This deterministic model has the same details as the model described in Section 3, except that the scene representation $c$ is not provided as input. The model only reasons about multi-agent interactions absent from scene context information.

\textbf{Multi Agent Scene}: The  deterministic model introduced in Section 3.

\textbf{GAN}: The  stochastic model introduced in Section 3.3. Similar to Social GAN~\cite{social_gan}, we sample $N$ times and report the best trajectory in the L2 sense for fair comparison with stochastic models, with $N = 3$ in Section 5.1, and $N = 20$ as adopted by~\cite{social_gan} in Section 5.2.

See \textbf{Supplementary Materials} for implementation details.

\section{Results}

\subsection{Driving Datasets}

\textbf{NGSIM Dataset}. We adopt the same experimental setting and directly report the presented results as in~\cite{conv_social_pooling}: We split the trajectories into segments of 8s, and all agents appearing in the 640-meter span are considered in the reasoning and prediction process. We use 3s of trajectory history and a 5s prediction horizon. LSTMs operate at 0.2s. As in~\cite{conv_social_pooling}, we report the Root Mean Square Error in meters with respect to each timestep $t$ within the prediction horizon: $ RMSE(t) = \sqrt { \frac {1}{n} \sum_ {i = 1, 2, .., n} ((\hat x_{it} - x_{it}) ^ 2 + (\hat y_{it} - y_{it})^2) } $
, where $n$ is the total number of agents in the validation set, $x_{it}$ denotes the $x$ coordinate of the $i$-th car in the dataset at future timestep $t$, and $y_{it}$ the $y$ coordinate at $t$.

\begin{table}[h!]
\small
\begin{center}
\begin{tabular}{l|c|c|c|c|c}
\textbf{Method} &  1s & 2s & 3s & 4s & 5s \\
\hline\hline
CV \cite{conv_social_pooling} & 0.73 & 1.78 & 3.13 & 4.78 & 6.68 \\
LSTM Baseline & 0.66 & 1.62 & 2.94 & 4.63 & 6.63 \\
C-VGMM + VIM \cite{vehicle} & \textbf{0.66} & 1.56 & 2.75 & 4.24 & 5.99 \\
\textbf{MATF Multi Agent} & 0.67 & \textbf{1.51} & \textbf{2.51} & \textbf{3.71} & \textbf{5.12} \\
\hline\hline
GAIL-GRU \cite{gail_gru} & 0.69 & 1.51 & 2.55 & 3.65 & 4.71 \\
\hline\hline
Social Conv \cite{conv_social_pooling} & \textbf{0.61} & \textbf{1.27} & 2.09 & 3.10 & 4.37 \\
\textbf{MATF GAN} & 0.66 & 1.34 & \textbf{2.08} & \textbf{2.97} & \textbf{4.13} \\
\hline\hline
\end{tabular}
\end{center}
\caption{Quantitative results on NGSIM~\cite{ngsim} dataset. RMSEs in meters with respect to each future timestep in the prediction horizon are reported.}
\end{table}

\begin{figure*}[ht]
   \includegraphics[width=\linewidth]{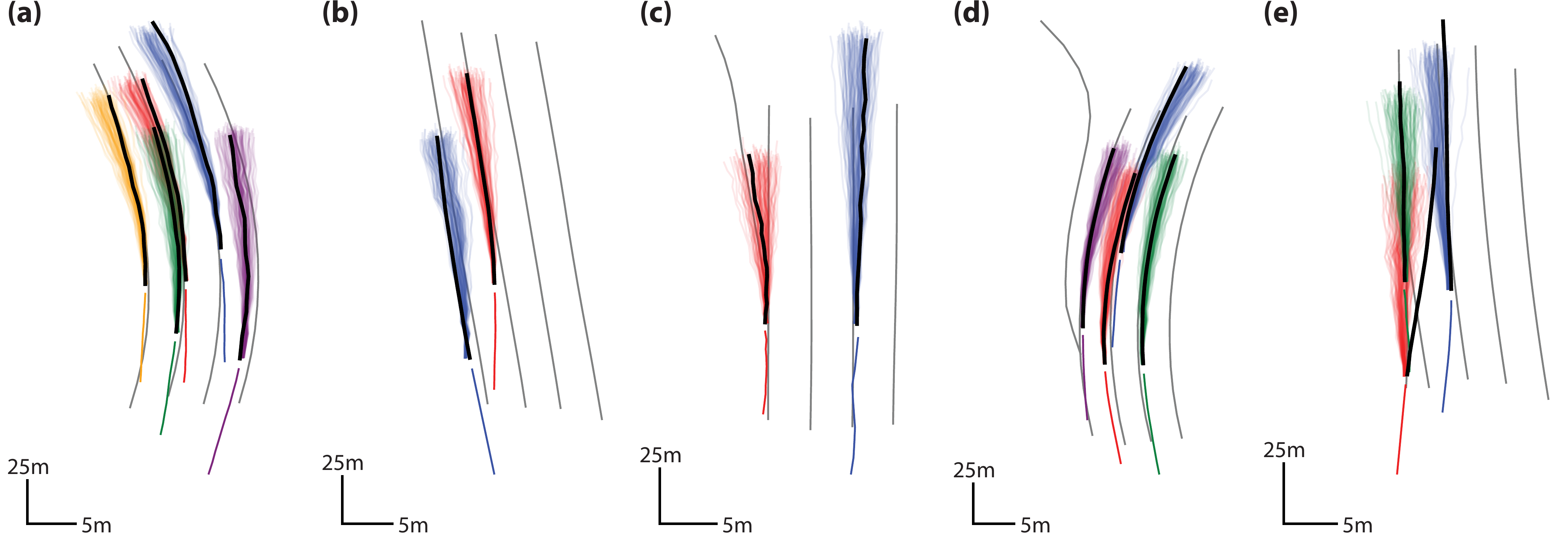}

   \caption{Qualitative results from Massachusetts driving dataset. Past trajectories are shown in different colors for each vehicle, followed by 100 sampled future trajectories. Ground truth future trajectories are shown in black, and lane centers are shown in gray.
   \textbf{(a)} A complex scenario involving five vehicles; MATF accurately predicts the trajectory and velocity profile for all.
   \textbf{(b)} MATF correctly predicts that the red vehicle will complete a lane change.
   \textbf{(c)} MATF captures the uncertainty over whether the red vehicle will take the highway exit.
   \textbf{(d)} As soon as the purple vehicle passes a highway exit, MATF predicts it will not take that exit.
   \textbf{(e)} Here, MATF fails to predict the precise ground truth trajectory; however, the red vehicle is predicted to initiate a lane change maneuver in a very small number of sampled trajectories.}
\label{fig:short}
\end{figure*}

Quantitative results are shown in Table 1. Our deterministic model \textit{MATF Multi Agent} outperforms the state-of-the-art deterministic model \textit{C-VGMM + VIM} \cite{vehicle}, a recent vehicle interaction approach based on variational Gaussian mixture models with Markov random fields. We include a comparison with \textit{GAIL-GRU}~\cite{gail_gru}; however, note that this model has access to the future ground-truth trajectories of other agents when predicting a given agent, while MATF and other models do not, so these results are not fully comparable. We compare our stochastic model, \textit{MATF GAN}, with \textit{Social Conv}~\cite{conv_social_pooling}, an approach that captures the distribution over future trajectories by representing maneuvers. \textit{MATF GAN} performs at the state-of-the-art level, with particularly improved performance at longer prediction horizons (3-5s). Note that \textit{Social Conv} has access to auxiliary supervision from maneuver labels, while MATF does not require this information. \textit{Multi Agent Scene} does not outperform \textit{Multi Agent} on NGSIM, because lanes in the NGSIM dataset are quite straight, and little agent-scene interaction is observed.

\textbf{Massachusetts Driving Dataset}. We also analyze a private Massachusetts driving dataset, which includes more curved lanes and more complex static scene contexts than NGSIM. NGSIM contains rich vehicle-vehicle interactions. However, the recorded highway span is quite straight, so minimal agent-scene interaction is observed. As an alternative, we analyze a large-scale dataset gathered during highway driving, including a several-mile stretch of highway, with curved lanes, highway exits, and entrances. Ablative studies are conducted for this dataset to show our model's ability to model agent-scene and agent-agent interactions, respectively. We report the Mean Absolute Error in meters with respect to each timestep $t$ within the prediction horizon: $ MAE(t) = \frac {1}{n} \sum_{i = 1, 2, .., n} \sqrt{(\hat x_{it} - x_{it}) ^ 2 + (\hat y_{it} - y_{it})^2}$.

\begin{figure}[ht]
\begin{center}
    \includegraphics[width=0.95\linewidth]{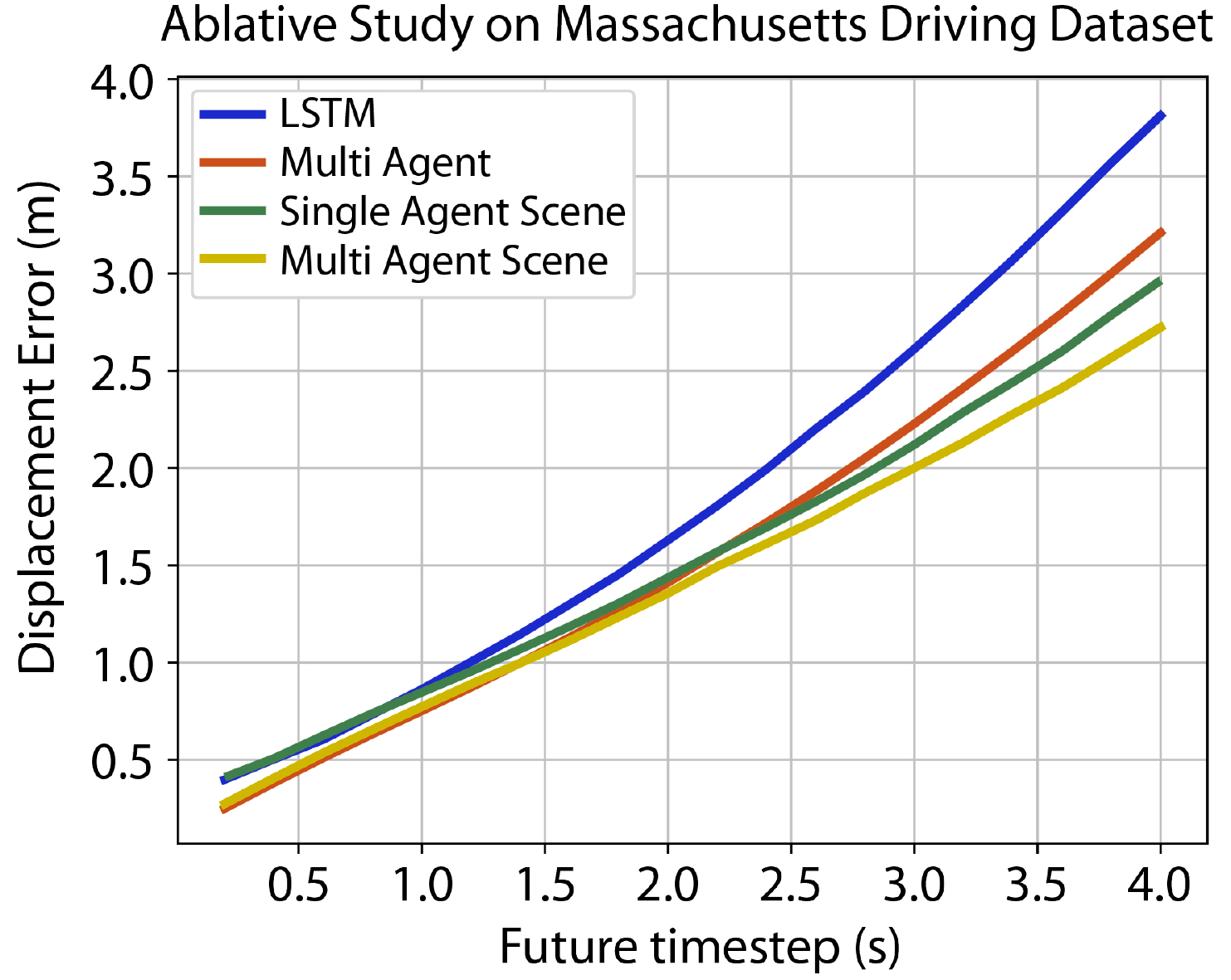}
\end{center}
   \caption{Quantitative results on Massachusetts driving dataset. MAEs in meters w.r.t. each future timestep in the prediction horizon are reported. Blue line is the evaluation result of \textit{LSTM}, red for \textit{Multi Agent}, green for \textit{Single Agent Scene}, and yellow for \textit{Multi Agent Scene}. Standard error around displacement error is plotted.}
\label{fig:short}
\end{figure}

Interesting qualitative results are shown in Fig.~3, and quantitative ablative results are shown in Fig.~4. Quantitative results show that both \textit{Single Agent Scene} and \textit{Multi Agent} outperform the \textit{LSTM} baseline, and that \textit{Multi Agent Scene} consistently outperforms \textit{Single Agent Scene} and \textit{Multi Agent}, and comparison between \textit{Multi Agent} and \textit{Single Agent Scene} shows that the former performs better at short term trajectory prediction, while the latter performs better at long term prediction.

From these studies, we conclude that our MATF model successfully models agent-agent and agent-scene interaction. More specifically, the scene fusion model learns constraints from the scene context, and the multi-agent model learns multi-agent interaction.

\subsection{Pedestrian Datasets}

\textbf{ETH-UCY Dataset}. We adopt the same experimental setting, split and error measure as \textit{Social GAN}: We split the trajectories into segments of 8s. We use 3.2s of trajectory history and a 4.8s prediction horizon. LSTMs operate at 0.4s. We use a leave-one-out approach, training on 4 sets and testing on the remaining set. We adopt exactly the same experimental settings, splits and error measures as~\cite{social_gan}. As in~\cite{social_gan}, we report the Average Displacement Error and Final Displacement Error in pixels with respect to each time-step $t$ within the prediction horizon:

\footnotesize
\begin{eqnarray*}
ADE(i) &=& \frac {1}{T'} \sum_ {j = 1, 2, .., T'} \sqrt{(\hat x_{ij} - x_{ij}) ^ 2 + (\hat y_{ij} - y_{ij})^2} \\
ADE &=& \frac{1}{n} \sum _ {i = 1, 2, .., n} ADE(i) \\
FDE(i) &=&  \sqrt{(\hat x_{iT'} - x_{iT'}) ^ 2 + (\hat y_{iT'} - y_{iT'})^2} \\
FDE &=& \frac{1}{n} \sum_{i = 1, 2, .., n} FDE(i),
\end{eqnarray*}
\normalsize
where $n$ is the total number of agents in the validation set, $x_{ij}$ and $y_{ij}$ denote the coordinates of the $i$-th agent in the dataset at future timestep $j$, and $T'$ denotes the final future timestep. Table 2 shows our results. MATF performs the best both in deterministic and stochastic settings.

\begin{table}[h]
\footnotesize
\begin{center}
{\setlength{\tabcolsep}{.5em}
\begin{tabular}{l|c|c|c|c}

\textbf{Dataset} & \multicolumn{2}{|c}{\textbf{Deterministic}} 
& \multicolumn{2}{|c}{\textbf{Stochastic}}\\

 & \textbf{S-LSTM} & \textbf{MATF} & \textbf{S-GAN} & \textbf{MATF GAN} \\

\hline\hline
\textbf{ETH} & $\mathbf{1.09}$ / $\mathbf{2.35}$ & $1.33$ / $2.49$ & $\mathbf{0.81}$ / $\mathbf{1.52}$ & $1.01$ / $1.75$ \\

\textbf{HOTEL} & $0.79$ / $1.76$ & $\mathbf{0.51}$ / $\mathbf{0.95}$ & $0.67$ / $1.37$ & $\mathbf{0.43}$ / $\mathbf{0.80}$ \\

\textbf{UNIV} & $0.67$ / $1.40$ & $\mathbf{0.56}$ / $\mathbf{1.19}$ & $0.60$ / $1.26$ & $\mathbf{0.44}$ / $\mathbf{0.91}$ \\

\textbf{ZARA1} & $0.47$ / $1.00$ & $\mathbf{0.44}$ / $\mathbf{0.93}$ & $0.34$ / $0.68$ & $\mathbf{0.26}$ / $\mathbf{0.45}$ \\

\textbf{ZARA2} & $0.56$ / $1.17$ & $\mathbf{0.34}$ / $\mathbf{0.73}$ & $0.42$ / $0.84$ & $\mathbf{0.26}$ / $\mathbf{0.57}$ \\

\hline
\textbf{AVG} & $0.72$ / $1.54$ & $\mathbf{0.64}$ / $\mathbf{1.26}$ & $0.57$ / $1.13$ & $\mathbf{0.48}$ / $\mathbf{0.90}$ \\

\hline\hline

\end{tabular}
}
\end{center}
\caption{Quantitative results on ETH-UCY datasets.  ADE / FDE of world coordinates in meters at $4.8s$ prediction horizon are reported. Our deterministic \textit{MATF} model outperforms \textit{Social LSTM}, and our stochastic \textit{MATF GAN} outperforms \textit{Social GAN}. We directly report the \textit{Social LSTM} and \textit{Social GAN} results presented in~\cite{social_gan}.}
\end{table}

\begin{figure*}[h!]
\begin{center}
    \includegraphics[width=0.309\linewidth]{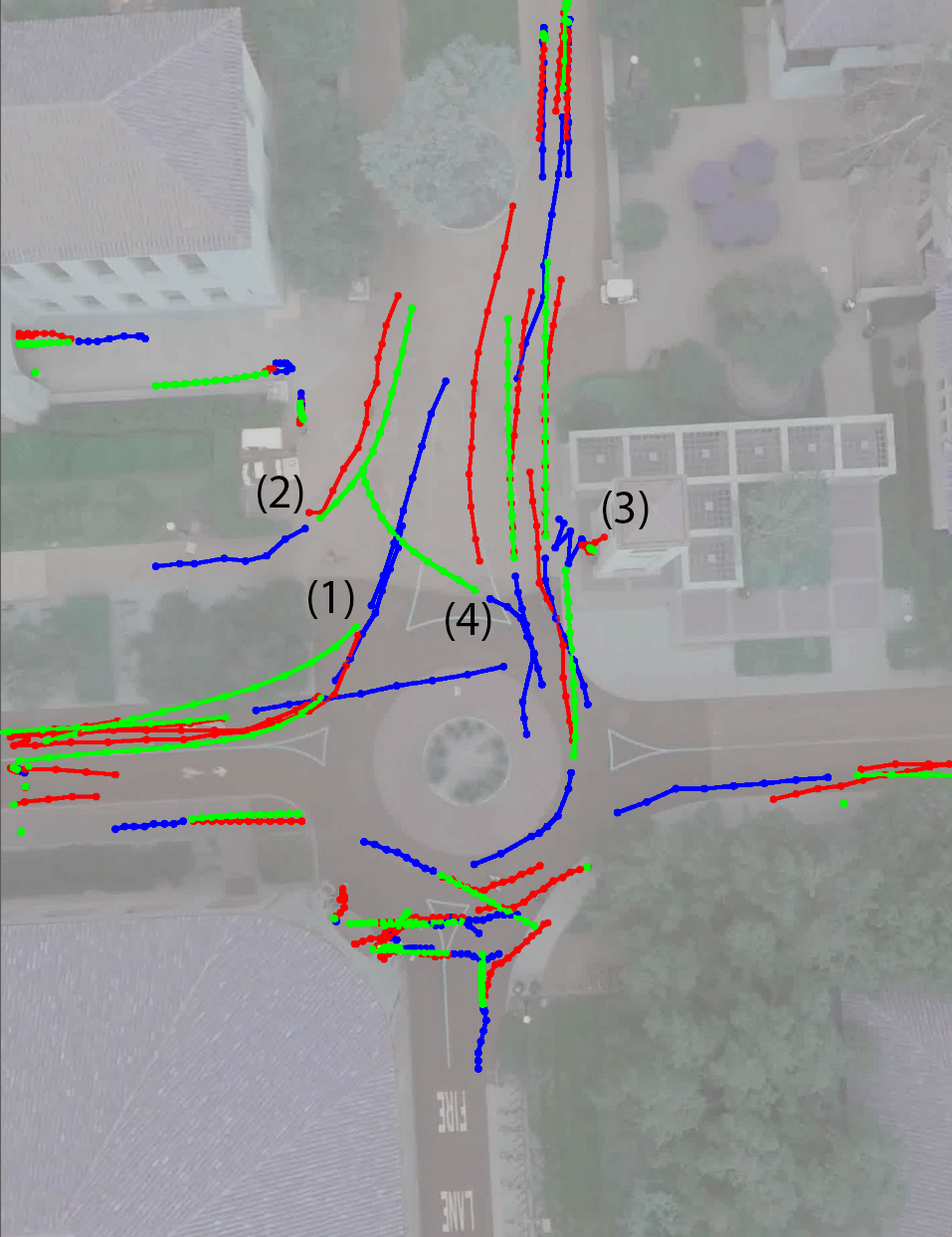}
    \includegraphics[width=0.303\linewidth]{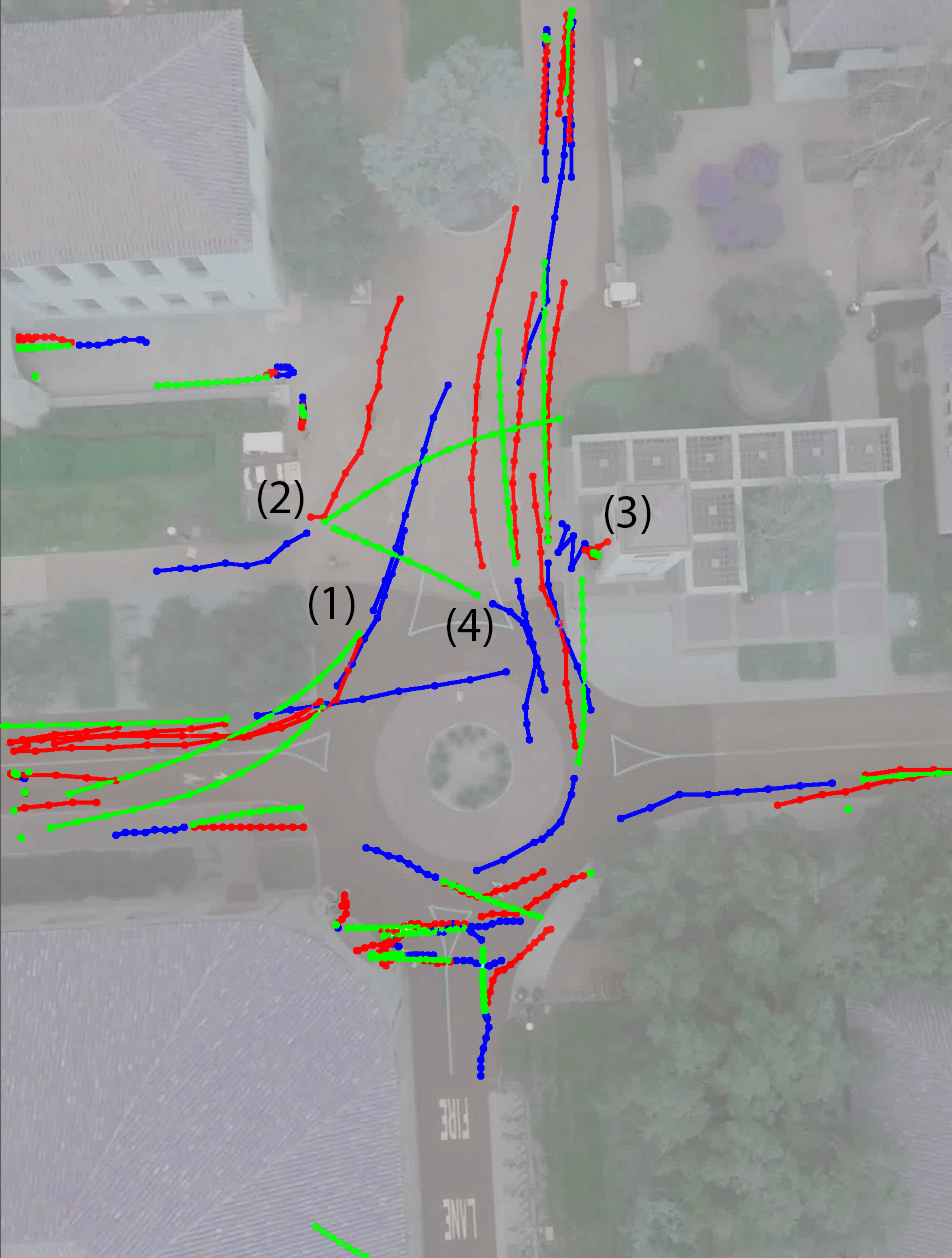}
    \includegraphics[width=0.3045\linewidth]{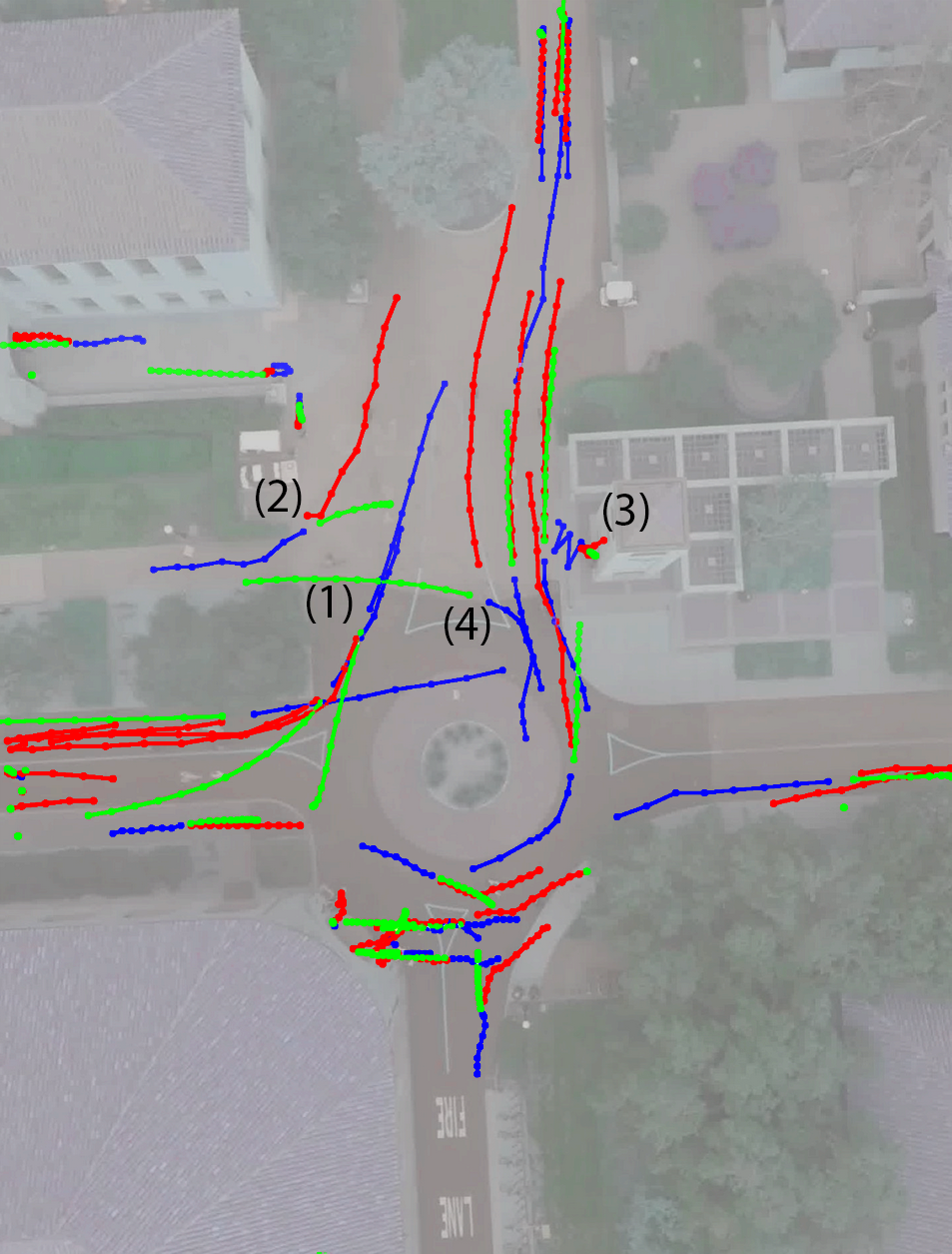}
\end{center}
   \caption{Ablative results on Stanford Drone dataset. From left to right are results from \textit{MATF Multi Agent Scene}, \textit{MATF Multi Agent}, and \textit{LSTM}, all deterministic models. Blue lines show past trajectories, red ground truth, and green predicted. All results come from the qualitative validation dataset. All the agent trajectories shown in this figure are predicted jointly via one forward pass. The closer the green predicted trajectory is to the red ground truth future trajectory, the more accurate the prediction. Our model predicts that (1) two agents entering the roundabout from the top will exit to the left; (2) one agent coming from the left on the pathway above the roundabout is turning left to move toward the top of the image; (3) one agent is decelerating at the door of the building above and to the right of the roundabout. (4) In one interesting failure case, an agent on the top-right of the roundabout is turning right to move toward  the top of the image; the model predicts the turn, but not how sharp it will be. 
These and various other qualitative patterns are correctly predicted by our Multi Agent Scene model, and some of them are approximated by our Multi Agent model, but most are not predicted by the baseline LSTM model.}
\label{fig:short}
\end{figure*}

\textbf{Stanford Drone Dataset}. We adopt the same experimental setting and directly report the results  presented in~\cite{sophie}: We split the trajectories into segments of 8s, and all agents appearing in the scene are considered in the reasoning and prediction process. We use 3.2s of trajectory history and a 4.8s prediction horizon. LSTMs operate at 0.4s per timestep. As in~\cite{stanford_drone}, we report ADE and FDE.

Fig. 5 shows qualitative ablative results using deterministic models; only the full \textit{MATF Multi Agent Scene} model captures the range of behaviors in the data.
Quantitative results for deterministic and stochastic models are shown in Table 3. \textit{MATF Multi Agent Scene} outperforms other deterministic models in ADE, and  \textit{MATF GAN} performs close to the state-of-the-art level. Among the deterministic models, \textit{Social LSTM} achieves the best performance in FDE. Among the stochastic models, \textit{Desire} gains strength from using Variational Auto-Encoders~\cite{vae} and Inverse Optimal Control to generate and rank trajectories; \textit{Sophie} performs the best with its strong attention-based social and physical reasoning modules. However, the computational complexity of these approaches is higher than that of other approaches due to the iterative process of IOC and $O(n^2)$-based attention mechanisms, respectively. In contrast, our model is more efficient in computational complexity with our shared convolution operations.

\begin{table}[h!]
\begin{center}
\footnotesize
\begin{tabular}{c|l|c|c|c}
\multicolumn{2}{c|}{Method} & ADE & FDE & Complexity \\
\hline\hline
\multirow{5}{*}{\rotatebox[origin=c]{90}{Deterministic}} & LSTM Baseline & 37.35 & 77.13 & $O(n)$ \\
~ & Social Force \cite{social_force} & 36.38 & 58.14 & $O(n)$ \\
~ & Social LSTM \cite{social_lstm} & 31.19 & \textbf{56.97} & $O(n)$ \\
~ & \textbf{MATF Multi Agent} & 30.75 & 65.90 & $O(n)$ \\
~ & \textbf{MATF Multi Agent Scene} & \textbf{27.82} & 59.31 & $O(n)$ \\
\hline\hline
\multirow{4}{*}{\rotatebox[origin=c]{90}{Stochastic}} & Social GAN \cite{social_gan} & 27.25 & 41.44 & $O(n)$ \\
~ & Desire \cite{desire} & 19.25 & 34.05 & $O(nK)$ \\
~ & Sophie \cite{sophie} & \textbf{16.27} & \textbf{29.38} & $O(n^2)$ \\
~ & \textbf{MATF GAN} & \textbf{22.59} & \textbf{33.53} & $O(n)$ \\
\hline\hline
\end{tabular}
\end{center}
\caption{Quantitative results on Stanford Drone \cite{stanford_drone} dataset. Average and Final Displacement Errors are reported. Computational complexity w.r.t agents number $n$ in a given scene is presented.}
\end{table}

We also analyze the factors influencing performance in our model---particularly the impact of the spatial resolution of the Multi-Agent Tensor. Table 4 shows that there is a U-shaped performance curve due to under/overfitting at low/high resolution, respectively, and that the ideal resolution is $32\times 32$, the setting we report. 

\begin{table}[h!]
\footnotesize
\begin{center}
{\setlength{\tabcolsep}{.5em}
\begin{tabular}{l|c|c|c|c|c|c}

\multicolumn{2}{c|}{\textbf{Spatial Grid Resolution}} & $4^2$ & $8^2$ & $16^2$ & $32^2$ & $64^2$ \\

\hline\hline
\textbf{Deterministic} &
\textbf{ADE} & $32.08$ & $32.36$ & $30.26$ & $\mathbf{27.82}$ & $29.47$  \\

& \textbf{FDE} & $68.08$ & $66.46$ & $62.73$ & $\mathbf{59.31}$ & $62.60$  \\

\hline

\textbf{Stochastic} &
\textbf{ADE} & $24.57$ & $23.55$ & $22.69$ & $\mathbf{22.59}$ & $23.50$  \\

& \textbf{FDE} & $39.44$ & $36.46$ & $\mathbf{33.45}$ & $33.53$ & $35.72$  \\

\hline\hline

\end{tabular}
}
\end{center}
\caption{Effect of spatial grid resolution on prediction accuracy. Results reported on Stanford Drone Dataset of $4.8s$ horizon.}
\end{table}

\section{Discussion}

We proposed an architecture for trajectory prediction which models scene context constraints and social interaction while retaining the spatial structure of multiple agents and the scene, unlike the purely agent-centric approaches more commonly used in the literature. Our motivation was that scene context constraints and social interaction patterns are invariant to the absolute coordinates where they take place; these patterns only depend on the relative positions among agents and scenes.
Convolutional layers are suited to modeling these kinds of position-invariant spatial interactions by sharing parameters across agents and space, while recent approaches like Social Pooling~\cite{social_lstm, social_gan} or Attention mechanisms~\cite{social_attention} cannot explicitly reason about spatial relationships among agents and cannot reason about these relationships at multiple spatial scales. Our Multi-Agent tensor fusion architecture models this naturally. To the best of our knowledge, MATF is the first approach which fuses information from a static scene context with multiple dynamic agent states, while retaining their spatial structure throughout the reasoning process to bridge the gap between agent-centric and spatial-centric trajectory prediction paradigms.

We applied our model to two different trajectory prediction tasks to demonstrate its flexibility and capacity to learn different types of behaviors, agent types, and scenarios from data. In the vehicle prediction domain, our model achieved state-of-the-art results at long-range prediction of vehicle trajectories in the NGSIM dataset. Our adversarially trained stochastic prediction model performed best relative to the maneuver-based approach of~\cite{conv_social_pooling}, suggesting that a representation of the distribution over maneuvers was necessary -- whether explicit as in~\cite{conv_social_pooling} or implicit as in our work. Our ablative studies on a Massachusetts driving dataset showed that representations of both the scene and multiagent interactions were necessary for accurate trajectory prediction in more complex scene contexts than NGSIM (greater lane curvature, more entrances and exits, etc.).

Our application to a state-of-the-art pedestrian dataset~\cite{stanford_drone} demonstrated comparable performance with previously published results. Although some recent models achieved  greater accuracy than ours~\cite{sophie,desire}, all used dramatically different architectures; it is interesting to find that a novel spatial-centric architecture can also achieve a high standard of performance. Future work should examine the factors that influence performance, and the advantages and disadvantages of different architectures.

In future work, we plan to integrate unsupervised learning of structured maneuver representations into our framework. This will increase the interpretability of our model predictions, while enabling our model to better capture multimodal structure in the distribution over agent-scene and agent-agent interactions.

Social trajectory prediction is a complex task, which depends on the ability to extract structure from the scene and the history of agents' joint motions. Our central goal here has been to combine the strengths of agent- and spatial-centric approaches to this problem. Beyond achieving more accurate multi-agent trajectory predictions, our belief is that the work of engineering better models will continue to yield further insights into the structure of human interaction.

\section{Acknowledgements}
This  work  was  mainly  conducted  at  ISEE,  Inc. with the support of the ISEE team and ISEE data platform. This work was supported in part by NSFC-61625201, 61527804.

{\small
\bibliographystyle{ieee}
\bibliography{egbib.bib}
}

\end{document}